\documentclass[runningheads]{llncs}
\usepackage{graphicx}
\usepackage[utf8]{inputenc} 
\usepackage[T1]{fontenc}    
\usepackage{hyperref}       
\usepackage{url}            
\usepackage{booktabs}       
\usepackage{amsfonts}       
\usepackage{nicefrac}       
\usepackage{microtype}      
\usepackage{xcolor}         

\usepackage{rotating}
\usepackage{amsmath,amsfonts,amscd,amssymb}
\usepackage{mathtools}
\usepackage{dsfont}
\usepackage{tikz}
\usepackage[mathscr]{euscript}
\usepackage[noend]{algorithmic}
\usepackage{algorithm}
\usepackage{setspace}
\usepackage{lineno}
\usepackage{todonotes}
\usepackage{comment}
\usetikzlibrary{decorations.pathreplacing}
\definecolor{color1bg}{HTML}{f73d28}
\definecolor{color2bg}{HTML}{FA8072}
\definecolor{bblue}{HTML}{00BFFF}
\definecolor{bblue2}{HTML}{00ffff}
\DeclareMathOperator*{\argminA}{arg\,min}

\usetikzlibrary{arrows,calc}
\tikzset{
>=stealth',
help lines/.style={dashed, thick},
axis/.style={<->},
important line/.style={thick},
connection/.style={thick, dotted},
}

\usetikzlibrary{shadows}
\usetikzlibrary{backgrounds}
\tikzset{
diagonal fill/.style 2 args={fill=#2, path picture={
		\fill[#1, sharp corners] (path picture bounding box.south west) -|
		(path picture bounding box.north east) -- cycle;}},
reversed diagonal fill/.style 2 args={fill=#2, path picture={
		\fill[#1, sharp corners] (path picture bounding box.north west) |- 
		(path picture bounding box.south east) -- cycle;}}
}
\usetikzlibrary{arrows.meta}

\begin{document}
\title{The Lattice Overparametrization Paradigm for the Machine Learning of Lattice Operators\thanks{Corresponding author: D. Marcondes (\email{dmarcondes@ime.usp.br}). D. Marcondes was funded by grants \#22/06211-2 and \#23/00256-7, São Paulo Research Foundation (FAPESP), and J. Barrera was funded by grants \#14/50937-1 and \#2020/06950-4, São Paulo Research Foundation (FAPESP).}}
\titlerunning{Lattice Overparametrization Paradigm}
%
\author{Diego Marcondes\inst{1,2}\orcidID{0000-0002-6087-4821} \and
	Junior Barrera\inst{1}\orcidID{0000-0003-0439-0475}}
\authorrunning{D. Marcondes and J. Barrera}
%
\institute{Department of Computer Science, Institute of Mathematics and Statistics, University of São Paulo, São Paulo, Brazil.  \and
	Department of Electrical and Computer Engineering, Texas A\&M University, College Station, USA}
\maketitle              
\begin{abstract}
	The machine learning of lattice operators has three possible bottlenecks. From a statistical standpoint, it is necessary to design a constrained class of operators based on prior information with low bias, and low complexity relative to the sample size. From a computational perspective, there should be an efficient algorithm to minimize an empirical error over the class. From an understanding point of view, the properties of the learned operator need to be derived, so its behavior can be theoretically understood. The statistical bottleneck can be overcome due to the rich literature about the representation of lattice operators, but there is no general learning algorithm for them. In this paper, we discuss a learning paradigm in which, by overparametrizing a class via elements in a lattice, an algorithm for minimizing functions in a lattice is applied to learn. We present the stochastic lattice descent algorithm as a general algorithm to learn on constrained classes of operators as long as a lattice overparametrization of it is fixed, and we discuss previous works which are proves of concept. Moreover, if there are algorithms to compute the basis of an operator from its overparametrization, then its properties can be deduced and the understanding bottleneck is also overcome. This learning paradigm has three properties that modern methods based on neural networks lack: control, transparency and interpretability. Nowadays, there is an increasing demand for methods with these characteristics, and we believe that mathematical morphology is in a unique position to supply them. The lattice overparametrization paradigm could be a missing piece for it to achieve its full potential within modern machine learning.
	
\end{abstract}

\section{Algebraic representations of operators}

Let $(\mathcal{L},\leq)$ be a complete lattice. A lattice operator $\psi: \mathscr{L} \to \mathscr{L}$ is a mapping from $\mathscr{L}$ into itself, and we denote by $\Psi = \mathcal{L}^{\mathscr{L}}$ the set of all lattice operators in $\mathcal{L}$. The collection $\Psi$ inherits the complete lattice structure of $\mathscr{L}$ by considering the pointwise partial order. Let $\Omega \subset \Psi$ be a complete sublattice of $\Psi$.

An algebraic representation of $(\Omega,\leq)$ is any complete lattice $(\Theta,\leq)$ such that there exists a lattice isomorphism $R: \Omega \to \Theta$. The element $\theta \in \Theta$ is the parameter that represents the operator $\psi_{\theta} = R^{-1}(\theta)$ and $(R,\Theta)$ is a parametrization of $\Omega$. The algebraic representations are not unique and, although they are all equivalent, some have advantages over others.

A general algebraic representation of a lattice operator $\psi$ is through its kernel, as proposed in\footnote{We are calling kernel what \cite{banon1993decomposition} defined as left-kernel.} \cite{banon1993decomposition}. Let $\Theta_{\mathcal{K}} = \mathcal{P}(\mathscr{L})^{\mathscr{L}}$ be the collection of all maps $\mathscr{F}$ from $\mathscr{L}$ to $\mathcal{P}(\mathscr{L})$ equipped with the pointwise partial order
\begin{linenomath}
	\begin{equation*}
		\mathscr{F}_{1} \leq \mathscr{F}_{2} \iff \mathscr{F}_{1}(Y) \subset \mathscr{F}_{2}(Y) \ \ \forall Y \in \mathscr{L}
	\end{equation*}
\end{linenomath}
for $\mathscr{F}_{1}, \mathscr{F}_{2} \in \Theta_{\mathcal{K}}$, and consider the lattice isomorphism $R_{\mathcal{K}}: \Omega \to \Theta_{\mathcal{K}}$ given by
\begin{linenomath}
	\begin{align*}
		R_{\mathcal{K}}(\psi)(Y) = \mathcal{K}(\psi)(Y) = \left\{X \in \mathscr{L}: Y \leq \psi(X)\right\} & & (Y \in \mathscr{L}).
	\end{align*}
\end{linenomath}
See \cite[Proposition~6.1]{banon1993decomposition} for a proof that $R_{\mathcal{K}}$ is a lattice isomorphism.

The operators in specific lattices, such as finite lattices, and subclasses of operators in general lattices, such as upper semi-continuous operators \cite{barrera1992expressiveness}, have a minimal algebraic representation by the maximal intervals lesser or equal to the kernel. Formally, for $\psi \in \Psi$ let
\begin{linenomath}
	\begin{equation*}
		\boldsymbol{A}(\mathcal{K}(\psi)) = \left\{[\alpha,\beta]: [\alpha,\beta] \leq \mathcal{K}(\psi)\right\}
	\end{equation*}
\end{linenomath}
be the intervals\footnote{See \cite{banon1993decomposition} for the formal definition of interval in this context.} which are lesser or equal to the kernel of $\psi$. The basis of $\psi$ is defined as the maximal intervals in $\boldsymbol{A}(\psi)$, that is
\begin{linenomath}
	\begin{align*}
		\boldsymbol{B}(\psi) &= \text{Max}\left(\boldsymbol{A}(\mathcal{K}(\psi))\right)\\
		&=\left\{[\alpha,\beta] \in \boldsymbol{A}(\psi): [\alpha^{\prime},\beta^{\prime}] \in \boldsymbol{A}(\psi),[\alpha,\beta] \leq [\alpha^{\prime},\beta^{\prime}] \implies [\alpha,\beta] = [\alpha^{\prime},\beta^{\prime}] \right\}
	\end{align*}
\end{linenomath}
in which $\leq$ above is the partial order in $(\Theta_{\mathcal{K}},\leq)$. Under certain conditions, of which more details may be found in \cite{banon1991minimal,banon1993decomposition,barrera1992expressiveness}, it follows that
\begin{linenomath}
	\begin{equation}
		\label{sup_gen}
		\psi = \vee \left\{\lambda_{[\alpha,\beta]} = \overline{\alpha} \wedge \overline{\beta}: [\alpha,\beta] \in \boldsymbol{B}(\psi)\right\}
	\end{equation}
\end{linenomath}
in which
\begin{linenomath}
	\begin{align*}
		\overline{\alpha}(X) = \vee \left\{Y \in \mathscr{L}: \alpha(Y) \leq X\right\} & & \text{ and } & & \overline{\beta}(X) = \vee \left\{Y \in \mathscr{L}: X \leq \beta(Y)\right\}
	\end{align*}
\end{linenomath}
for $X \in \mathscr{L}$. Decomposition \eqref{sup_gen} is called sup-generating decomposition of $\psi$ and it has a dual inf-generating decomposition.

Assuming that \eqref{sup_gen} holds for all $\psi \in \Omega$, denote by
\begin{linenomath}
	\begin{equation*}
		\Theta_{\boldsymbol{B}} = \left\{\text{Max}\left(\boldsymbol{A}(\mathscr{F})\right): \mathscr{F} \in \Theta_{\mathcal{K}}\right\}
	\end{equation*}
\end{linenomath}
the maximal intervals associated to each $\mathscr{F} \in \Theta_{\mathcal{K}}$ and consider the map $R_{\boldsymbol{B}}:\Theta_{\mathcal{K}} \to \Theta_{\boldsymbol{B}}$ given by
\begin{linenomath}
	\begin{equation*}
		R_{\boldsymbol{B}}(\mathscr{F}) = \text{Max}\left(\boldsymbol{A}(\mathscr{F})\right).
	\end{equation*}
\end{linenomath}
It follows that $(\Theta_{\boldsymbol{B}},\leq)$ is a complete lattice isomorphic to $(\Theta_{\mathcal{K}},\leq)$ with partial order
\begin{linenomath}
	\begin{equation*}
		\boldsymbol{B}_{1} \leq \boldsymbol{B}_{2} \iff \forall [\alpha,\beta] \in \boldsymbol{B}_{1}, \exists [\alpha^{\prime},\beta^{\prime}] \in \boldsymbol{B}_{2}: [\alpha,\beta] \leq [\alpha^{\prime},\beta^{\prime}]
	\end{equation*}
\end{linenomath}
in which the partial order on the right-hand side is that of $(\Theta_{\mathcal{K}},\leq)$. From now on, we assume that $\Omega$ is a subclass of operators on $\mathscr{L}$ with a basis representation.

Specific classes of operators may have other algebraic representations. For instance, when $\mathscr{L} = \mathcal{P}(E)$ and $(E,+)$ is an Abelian group, then the class of translation invariant (t.i.) and locally defined lattice operators (i.e., W-operators), which in this case are set operators, can also be represented by a characteristic Boolean function. Denoting by $\Psi_{W}$ the class of t.i. set operators locally defined within a window $W \in \mathcal{P}(E)$ and by $\mathfrak{B} = \{0,1\}^{\mathcal{P}(W)}$ the Boolean functions in $\mathcal{P}(W)$, we consider the lattice isomorphism $R_{\mathfrak{B}}: \Psi_{W} \to \mathfrak{B}$ given by
\begin{linenomath}
	\begin{align*}
		R_{\mathfrak{B}}(\psi)(X) = \begin{cases}
			1, & \text{ if } o \in X\\
			0, & \text{ otherwise}
		\end{cases} & & (X \in \mathcal{P}(W))
	\end{align*}
\end{linenomath}
in which $o$ is the zero element $E$. See \cite{barrera1996set} for more details.

Clearly, the isomorphisms may be composed to obtain isomorphisms between distinct algebraic representations and all algebraic representations are equivalent. For example, $R_{\mathfrak{B},\boldsymbol{B}}: \mathfrak{B} \to \boldsymbol{B}$ given by $R_{\mathfrak{B},\boldsymbol{B}} = R_{\boldsymbol{B}} \circ R_{\mathcal{K}} \circ R_{\mathfrak{B}}^{-1}$ is an isomorphism between $(\mathfrak{B},\leq)$ and $(\boldsymbol{B},\leq)$. The isomorphisms defined so far are illustrated in Figure \ref{lattice_iso}.

From an algebraic perspective, the basis representation has some advantages over other representations, since algebraic properties of an operator may be deduced from its basis. For example, in the case of W-operators, the intervals in the basis of increasing operators are of form $[A,W]$ for $A \in \mathcal{P}(W)$; the basis of extensive increasing operators contains the interval $[o,W]$; and the basis of an increasing anti-extensive operator is such that $o \in A$ for all lower limits $A$ of the intervals in its basis (see \cite{jones1994basis} for more details). Hence, reducing an operator to its basis representation is enough to verify its mathematical properties.

\section{Lattice overparametrization}

An algebraic representation $R$ is an isomorphism between a class $\Omega$ and a parametric set $\Theta$. Such a representation is obtained by departing from a fixed $\Omega$ and defining an isomorphism $R: \Omega \to \Theta$. This is done in \cite{banon1991minimal,banon1993decomposition,barrera1992expressiveness}. Another family of representations may be obtained by departing from a $\Theta$ and defining an onto map $\tilde{R}: \Theta \to \Omega$, so each parameter $\theta \in \Theta$ represents an operator $\psi_{\theta} = \tilde{R}(\theta) \in \Omega$ and for each $\psi \in \Omega$ there exists \textit{at least one} $\theta \in \Theta$ such that $\psi = \psi_{\theta}$. 

When $\tilde{R}$ is not injective, $(\tilde{R},\Theta)$ is an overparametrization of $\Omega$ by the parameters in $\Theta$ since a same operator can be represented by more than one parameter. If $(\Theta,\leq)$ is a lattice, we say that $(\tilde{R},\Theta)$ is a lattice overparametrization of $\Omega$. Since $\tilde{R}$ is not an isomorphism, the partial relation in $\Theta$ is not equivalent to that in $\Omega$. The basis of the operator represented by $\theta$ is given by $\tilde{R}_{\boldsymbol{B}}(\theta) = (R_{\boldsymbol{B}} \circ R_{\mathcal{K}} \circ \tilde{R})(\theta)$. 

As an example, assume that $\mathscr{L} = \mathcal{P}(E)$ and $(E,+)$ is an Abelian group. For a finite subset $W \in \mathcal{P}(E)$ let
\begin{linenomath}
	\begin{align}
		\label{constraint_class}
		\Omega = \left\{\epsilon_{A} \vee \epsilon_{B} \vee \epsilon_{C}: A,B,C \in \mathcal{P}(W);\{[A,W],[B,W],[C,W]\} \text{ is maximal}\right\}
	\end{align}
\end{linenomath}
be the class of t.i. operators locally defined within $W$ that can be written as the supremum of three erosions. We note that $\boldsymbol{B}(\epsilon_{A} \vee \epsilon_{B} \vee \epsilon_{C}) = \left\{[A,W],[B,W],[C,W]\right\}$ and $\Omega$ is actually the class of the increasing $W$-operators with at most three elements in their basis\footnote{Observe that if some of the elements $A,B,C$ are equal, then $|\boldsymbol{B}(\epsilon_{A} \vee \epsilon_{B} \vee \epsilon_{C})| < 3$.}. By making $\Theta = \mathcal{P}(W)^{3}$ and $\tilde{R}((A,B,C)) = \epsilon_{A} \vee \epsilon_{B} \vee \epsilon_{C}$ we have a lattice overparametrization of $\Omega$ since $\tilde{R}((A,B,C)) = \psi$ for all $(A,B,C) \in \mathcal{P}(W)^{3}$ satisfying $\mathcal{K}(\psi) = [A,W] \cup [B,W] \cup [C,W]$. By lifting the restriction that the intervals $\{[A,W],[B,W],[C,W]\}$ are maximal, we depart from an algebraic representation of $\Omega$ to a lattice overparametrization by a Boolean lattice.

A lattice overparametrization may be useful for representing a constrained class of operators defined via the composition, supremum, and infimum of operators that can be parametrized by elements in a lattice. A special case is when the operators can be written as combinations of erosions and dilations with structural elements in a lattice. In \cite{marcondes2023discrete} we proposed the discrete morphological neural networks (DMNN) to represent constrained classes of $W$-operators via the composition, supremum and infimum of W-operators, which are an example of overparametrizations of a class of operators. In special, the canonical DMNN are those in which the $W$-operators computed in the network can be written as the supremum, infimum, complement, or composition of erosions and dilations with a same structuring element, an example of which is the class in \eqref{constraint_class} (see Example 5.8 in \cite{marcondes2023discrete}). The canonical DMNN are a specific example of lattice overparametrization.

The main advantage of considering a lattice overparametrization is the possibility of applying general, efficient algorithms to learn operators in a constrained class. This is the case since the lattice $(\Theta,\leq)$ is known and can be chosen with desired computational properties, so minimizing a function in it may be more efficient than doing so in $(\Omega,\leq)$, specially when $\Omega$ is not a lattice. We further discuss the advantages of considering a lattice overparametrization to learn lattice operators in Section \ref{Sec4}.

\section{The machine learning of lattice operators}

The general framework for learning lattice operators consists of a class $\Omega$, a sample $\mathcal{D}_{N} = \{(X_{1},Y_{1}),\dots,(X_{N},Y_{N})\}$ of $N$ pairs of input and output elements $X$ and $Y$ in $\mathscr{L}$, in which $Y$ is obtained by a possibly random transformation of $X$, and a loss function $\ell: \mathscr{L}^{2} \times \Psi \to \mathbb{R}^{2}$ which evaluates the \textit{error} $\ell((X,Y),\psi)$ incurred when $\psi(X)$ is applied to approximate $Y$, for each pair $(X,Y) \in \mathscr{L}^{2}$ and operator $\psi \in \Psi$. 

It is assumed that the pairs in $\mathcal{D}_{N}$ are sampled from an unknown, but fixed, statistical distribution $P$ over $\mathscr{L}^{2}$. Each $\psi \in \Psi$ has a mean expected error under distribution $P$ defined as $L(\psi) = \mathbb{E}_{P}\left[\ell((X,Y),\psi)\right]$, in which the expectation is over a random vector $(X,Y)$ with distribution $P$. A target operator of $\Psi$ is a minimizer of $L$ in $\Psi$ and a target operator of $\Omega$ is a minimizer of $L$ in $\Omega$. We denote the target operators by $\psi^{\star}$ and $\psi^{\star}_{\Omega}$, respectively, and they satisfy $L(\psi^{\star}) \leq L(\psi), \forall \psi \in \Psi$, and $L(\psi^{\star}_{\Omega}) \leq L(\psi), \forall \psi \in \Omega$. For the sake of the argument, we assume that both target operators exist and are unique.

Defining
\begin{linenomath}
	\begin{equation*}
		L_{\mathcal{D}_{N}}(\psi) = \frac{1}{N} \sum_{i=1}^{N} \ell((X_{i},Y_{i}),\psi)
	\end{equation*}
\end{linenomath}
as the mean empirical error of $\psi \in \Psi$ in sample $\mathcal{D}_{N}$, the empirical risk minimization paradigm propose as an estimator for $\psi^{\star}_{\Omega}$ the operator that minimizes $L_{\mathcal{D}_{N}}$ in $\Omega$:
\begin{linenomath}
	\begin{equation}
		\label{psiHat}
		\hat{\psi} = \argminA\limits_{\psi \in \Omega} L_{\mathcal{D}_{N}}(\psi) = \argminA\limits_{\theta \in \Theta} L_{\mathcal{D}_{N}}(\psi_{\theta})
	\end{equation}
\end{linenomath}
in which $\Theta$ is any representation, algebraic or otherwise, of $\Omega$. The quality of the estimator $\hat{\psi}$ is measured by $L(\hat{\psi})$, which is called its generalization error, and assesses how it is expected to perform on data not in the sample, but generated by the same unknown distribution $P$.

The goal of learning is to obtain an estimator such that $L(\hat{\psi}) \approx L(\psi^{\star})$ so its generalization quality is close to the best possible. On the one hand, it is necessary to have $L(\psi^{\star}_{\Omega}) \approx L(\psi^{\star})$ for otherwise there is a systematic bias in the learning process since $\hat{\psi}$ cannot generalize better than $\psi^{\star}_{\Omega}$. On the other hand, if $\Omega$ is chosen as a class of complex operators, or as $\Omega = \Psi$, then, even if $\psi^{\star}_{\Omega}$ is as good as or equal to $\psi^{\star}$, if the sample size is not great enough, there may be a complex operator $\hat{\psi}$ in $\Omega$ that completely fits the data, so it has zero empirical error, but that does not generalize very well. When this happens, we say overfitting occurred. Actually, if $\Omega = \Psi$ and $\Psi$ has infinite VC dimension, which is a measure of the complexity of a class of operators \cite{vapnik1999nature}, not even an infinite sample suffices to guarantee that $L(\hat{\psi}) \approx L(\psi^{\star})$. This is the usual bias-variance trade-off in machine learning \cite{abu2012learning}.

Hence, we have the following statistical bottleneck for learning lattice operators:

\vspace{0.1cm}

\textbf{(B1)} \textit{To fix a class of operators with low bias and relative low complexity}

\vspace{0.1cm}

The recipe to circumvent \textbf{(B1)} is the core of mathematical morphology: to design a class of operators based on prior information about the practical problem and on the mathematical properties of lattice operators. Geometrical and topological properties of the transformation applied to $X_{i}$ to obtain $Y_{i}$ in sample $\mathcal{D}_{N}$ are identified, and based on them a class $\Omega$ of lattice operators is designed via the mathematical morphology toolbox. If prior information is right, so the best operator in $\Omega$ well generalizes, and $\Omega$ is not too complex, then learning is feasible and $\hat{\psi}$ is expected to well generalize. As an example, the class in \eqref{constraint_class} can be applied to a problem in which it is known that an increasing transformation was applied to $X_{i}$ to obtain $Y_{i}$, and the maximum number of elements in the basis controls the complexity of $\Omega$.

There are almost 60 years of rich literature in mathematical morphology, that we could not possibly cite here without committing huge injustices, which can be directly applied to solving \textbf{(B1)}, so it is not really a bottleneck for learning lattice operators. However, there is a second, computational, bottleneck that has not yet been overcome in general:

\vspace{0.1cm}

\textbf{(B2)} \textit{To compute $\hat{\psi}$ by solving \eqref{psiHat}}

\vspace{0.1cm}

Despite their practical success, many proposed methods for the machine learning of lattice operators in the literature are heuristics that seek to control the complexity of the class of operators relative to the sample size, but do not strongly restrict the operator class based on prior information. The ISI algorithm \cite{hirata2002incremental}, iterative designs \cite{hirata2000iterative} and multiresolution designs \cite{dougherty2001multiresolution,hirata2002multiresolution} offer methods to control the complexity of the class based on data, however are not flexible to represent specific classes of operators, but only general classes such as filters.

Furthermore, methods such as the those based on envelope constraints \cite{brun2003design,brun2004nonlinear} can insert sharp prior information into the learning process by projecting the operator learned by a heuristic method into a constrained class, but do not guarantee that the projected operator well approximates the target of the class. Finally, we note that methods to solve \eqref{psiHat} for specific classes, such as stack filters \cite{hirata1999design}, have been proposed, but are not general methods that can be easily extended to other classes of operators. See \cite{barrera2022mathematical} for more details on methods for the machine learning of operators.

We propose as a general paradigm to overcoming \textbf{(B2)} the development of algorithms to efficiently minimize, or approximately minimize, a function in a lattice so \eqref{psiHat} can be at least approximately computed whenever $\Omega$ has a lattice overparametrization $(\Theta,\leq)$. Such an algorithm would be a general optimizer for learning operators once a subclass $\Omega$ and a lattice overparametrization for it is fixed. This abstract idea, which is behind the DMNN proposed in \cite{marcondes2023discrete}, can be a paradigm for the machine learning of lattice operators based on the stochastic lattice descent algorithm (SLDA). The general framework for the machine learning of lattice operators is depicted in Figure \ref{lattice_iso}.

\section{Stochastic lattice descent as a general learning algorithm}
\label{Sec4}

The U-curve algorithm was first proposed by \cite{u-curve1} for minimizing U-shaped functions in Boolean lattices, and was then improved by \cite{u-curve3,ucurveParallel,reis2018}. It has also been applied to solve other problems in mathematical morphology \cite{reis2013solving}. Inspired by this algorithm and by the success of stochastic gradient descent algorithms for minimizing overparametrized functions in $\mathbb{R}^{d}$, such as the regularized empirical error of a neural network, we propose the SLDA to learn operators in a class with lattice overparametrization $(\Theta,\leq)$.

Informally, the SLDA performs a greedy search of a lattice to minimize an empirical error. At each step, $n$ neighbors of an element are sampled and the empirical error on a fixed sample batch of the operator represented by each sampled neighbor is calculated. The algorithm jumps to the sampled neighbor with the least empirical error on the sample batch. The algorithm starts again from this new element, by sampling $n$ neighbors and calculating their empirical error on a new sample batch. This process goes on for a predetermined number of epochs. An epoch ends when all sample batches have been considered, and the algorithm returns the element visited at the end of an epoch with the least empirical error on the whole sample. We now formally define the SLDA.

For each $\theta \in \Theta$, let $N(\theta)$ be a \textit{neighborhood} of $\theta$ in $(\Theta,\leq)$. If $\Theta$ is countable, then $N(\theta)$ may be composed by the elements of $\Theta$ at distance one from $\theta$. When $\Theta$ is uncountable and $d(\theta,\theta^{\prime})$ is a distance measure, with $d(\theta,\theta^{\prime}) = \infty$ whenever $\theta \nleq \theta^{\prime}$ and $\theta^{\prime} \nleq \theta$, then one could consider $N(\theta) = \left\{\theta^{\prime}: d(\theta,\theta^{\prime}) < \delta\right\}$ for a fixed $\delta > 0$. Assume that, given $\theta$ and a constant $n$, there exists an algorithm which samples $n$ elements from $N(\theta)$. If $N(\theta)$ is a finite set, then the elements may be sampled uniformly, while if it is countable or uncountable then other statistical distributions should be considered.

The SLDA is formalized in Algorithm \ref{A2}. The initial point $\theta \in \Theta$, a batch size\footnote{We assume that $N/b$ is an integer to easy notation. If this is not the case, the last batch will contain less than $b$ points.} $b$, the number $n$ of neighbors to be sampled at each step, and the number of training epochs is fixed. The initial point is stored as the point with minimum empirical error visited so far. For each epoch, the sample $\mathcal{D}_{N}$ is randomly partitioned in $N/b$ batches $\{\tilde{\mathcal{D}}^{(1)}_{b},\dots,\tilde{\mathcal{D}}^{(N/b)}_{b}\}$. For each batch $\tilde{\mathcal{D}}^{(j)}_{b}$, $n$ neighbors of $\theta$ are sampled and $\theta$ is updated to a sampled neighbor with the least empirical error $L_{\tilde{\mathcal{D}}^{(j)}_{b}}$, that is calculated on the sample batch $\tilde{\mathcal{D}}^{(j)}_{b}$. Observe that $\theta$ is updated at each batch, so during an epoch, it is updated $N/b$ times. 

At the end of each epoch, the empirical error $L_{\mathcal{D}_{N}}(\theta)$ of $\theta$ on the whole sample $\mathcal{D}_{N}$ is compared with the error of the point with the least empirical error visited so far at the end of an epoch, and it is stored as this point if its empirical error is lesser. After the predetermined number of epochs, the algorithm returns the point with the least empirical error on the whole sample $\mathcal{D}_{N}$ visited at the end of an epoch. For finite lattices, if $b = N$ and $n$ is equal to the number of neighbors of $\theta$, i.e., $n = n(\theta) = |N(\theta)|$, then Algorithm \ref{A2} reduces to the (deterministic) lattice descent algorithm.

\begin{algorithm}[ht]
	\centering
	\caption{Stochastic lattice descent algorithm for learning lattice operators.}
	\label{A2}
	\begin{algorithmic}[1]
		\ENSURE $\theta \in \Theta, n, b, Epochs$		
		\STATE $L_{min} \gets L_{\mathcal{D}_{N}}(\psi_{\theta})$
		\STATE $\widehat{\theta} \gets \theta$ 
		\FOR{run $\in \{1,\dots,\text{Epochs}\}$}
		\STATE $\{\tilde{\mathcal{D}}^{(1)}_{b},\dots,\tilde{\mathcal{D}}^{(N/b)}_{b}\} \gets \text{SampleBatch}(\mathcal{D}_{N},b)$
		\FOR{$j \in \{1,\dots,N/b\}$}
		\STATE $\tilde{N}(\theta) \gets \text{SampleNeighbors}(\theta,n)$	
		\STATE $\theta \gets \theta^{\prime} \text{ s.t. } \ \theta^{\prime} \in \tilde{N}(\theta) \text{ and } L_{\tilde{\mathcal{D}}^{(j)}_{b}}(\psi_{\theta^\prime})  = \min\{L_{\tilde{\mathcal{D}}^{(j)}_{b}}(\psi_{\theta^{\prime\prime}}): \theta^{\prime\prime} \in \tilde{N}(\theta)\}$
		\ENDFOR
		\IF{$L_{\mathcal{D}_{N}}(\psi_{\theta}) < L_{min}$}
		\STATE $L_{min} \gets L_{\mathcal{D}_{N}}(\psi_{\theta})$
		\STATE $\widehat{\theta} \gets \theta$ 
		\ENDIF
		\ENDFOR
		\RETURN{$\widehat{\theta}$}
	\end{algorithmic}
\end{algorithm}

An implementation of Algorithm \ref{A2} for a finite lattice has been done in \cite{marcondes2023discrete} and good results were obtained in a simple binary image transformation problem. We note that in order for the algorithm to work for uncountable lattices, the statistical distribution applied to sample the neighbors should be chosen in a way to give a meaningful probability to chains in which the error decreases. The challenge of doing so is defining such a distribution without computing the error on the chains, what is computationally unfeasible. An implementation of the SLDA, or a modification of it, for uncountable lattices is currently an open problem.

We argue that, in general, it is not computationally feasible to apply the SLDA directly on lattice $(\Omega,\leq)$. On the one hand, since $(\Theta,\leq)$ is known a priori, for any $\theta \in \Theta$ the set $N(\theta)$ is known, so the complexity of sampling $n$ neighbors should be that of sampling from a known statistical distribution, which is usually very low. On the other hand, if the SLDA was applied directly on $(\Omega,\leq)$, fixed a $\psi \in \Omega$, the computation of its neighborhood in $(\Omega,\leq)$ would be problem-specific and could have a great complexity. Therefore, suboptimally minimizing the empirical error in $\Theta$ via the SLDA should be less computationally complex than doing so in $\Omega$. Furthermore, it is possible to learn on a poset $(\Omega,\leq)$ as long as it has a lattice overparametrization. In this case, minimizing the empirical error in $(\Omega,\leq)$ is a constrained optimization problem, while minimizing it in the lattice $(\Theta,\leq)$ is an unconstrained one which ought to be more efficiently solved.

We also note that the SLDA could be applied to the case in which $\Theta$ is a poset possibly contained in a lattice. In this case, the complexity of the algorithm could increase significantly due to the restrictions on $N(\theta)$. For example, sampling $n$ neighbors of an element in a Boolean lattice is trivial, while sampling $n$ neighbors which are also in a set of elements (the poset $\Theta$) may be quite complex, specially when $\Theta \cap N(\theta)$ needs to be computed. In other cases, $\Theta$ being a poset may not meaningfully increase the complexity (see the application in \cite{marcondes2023discrete}).

\section{Degrees of prior information and hierarchical SLDA}

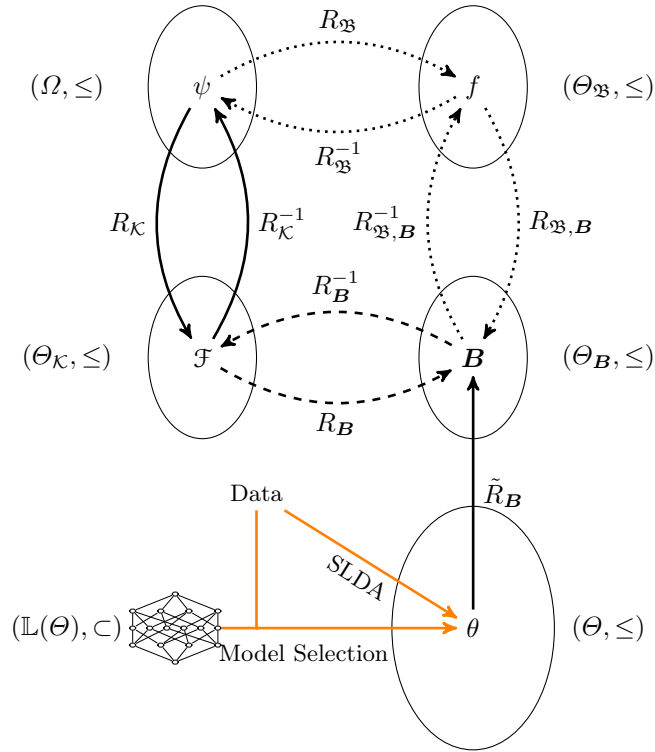
\begin{figure}[ht]
	\centering
	\begin{tikzpicture}[scale=0.72]
		\tikzstyle{hs} = [circle,draw=black, rounded corners,minimum width=3em, vertex distance=2.5cm, line width=1pt]
		\tikzstyle{hs2} = [circle,draw=black,dashed, rounded corners,minimum width=3em, vertex distance=2.5cm, line width=1pt]
		
		\draw (0,0) ellipse (1cm and 1.5cm);
		\draw (5,0) ellipse (1cm and 1.5cm);
		\draw (0,-5) ellipse (1cm and 1.5cm);
		\draw (5,-5) ellipse (1cm and 1.5cm);
		\draw (5,-10) ellipse (1.5cm and 2.25cm);
		
		\node (psi) at (0,0) {$\psi$};
		\node (f) at (5,0) {$f$};
		\node (IntX) at (5,-5) {$\boldsymbol{B}$};
		\node (X) at (0,-5) {$\mathscr{F}$};
		\node (theta) at (5,-10) {$\theta$};
		\node[inner sep=0pt] (LS) at (-0.5,-10) {\includegraphics[width=0.1\textwidth]{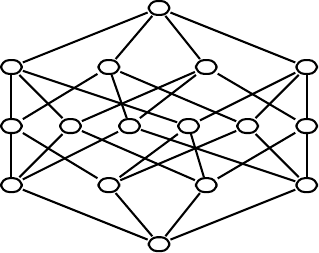}};
		\node (data) at (1,-7.5) {\footnotesize Data};
		\node (mid2) at (2.25,-10.5) {{\begin{tabular}{c} \footnotesize Model Selection\space \space \space \space \space \space\\ \end{tabular}}};
		\node (mid1) at (1,-10.2) {};

		\node at (-2.5,0) {$(\Omega,\leq)$};
		\node at (7.5,0) {$(\Theta_{\mathfrak{B}},\leq)$};
		\node at (7.5,-5) {$(\Theta_{\boldsymbol{B}},\leq)$};
		\node at (7.5,-10) {$(\Theta,\leq)$};
		\node at (-2.5,-5) {$(\Theta_{\mathcal{K}},\leq)$};
		\node at (-2.5,-10) {$(\mathbb{L}(\Theta),\subset)$};

		\begin{scope}[line width=1pt]
			\draw[->,dotted] (psi) to[bend left]  node[sloped,above] {$R_{\mathfrak{B}}$} (f);
			\draw[->,dotted] (f) to[bend left]  node[sloped,below] {$R_{\mathfrak{B}}^{-1}$} (psi);
			
			\draw[->] (psi) to[bend right]  node[left] {$R_{\mathcal{K}}$} (X);
			\draw[->] (X) to[bend right]  node[right] {$R_{\mathcal{K}}^{-1}$} (psi);
			
			\draw[->,dotted] (f) to[bend left]  node[right] {$R_{\mathfrak{B},\boldsymbol{B}}$} (IntX);
			\draw[->,dotted] (IntX) to[bend left]  node[left] {$R_{\mathfrak{B},\boldsymbol{B}}^{-1}$} (f);
			
			\draw[->,dashed] (X) to[bend right]  node[sloped,below] {$R_{\boldsymbol{B}}$} (IntX);
			\draw[->,dashed] (IntX) to[bend right]  node[sloped,above] {$R_{\boldsymbol{B}}^{-1}$} (X);
			
			\draw[->] (theta) to  node[right] {$\tilde{R}_{\boldsymbol{B}}$} (IntX);
			
			\draw[->,orange] (LS) to  (theta);
			\draw[-,orange] (data) to  (mid1);
			\draw[->,orange] (data) to node[sloped,below,black] {\footnotesize SLDA \space} (theta);
		\end{scope}
	\end{tikzpicture}
	\caption{\footnotesize The lattice isomorphisms between representations of $(\Omega,\leq)$. The dashed lines represent an isomorphism that holds when the operators in $\Omega$ have a basis representation. The dotted lines represent isomorphisms that hold for t.i. and locally defined set operators when $\mathscr{L} = \mathcal{P}(E)$ and $(E,+)$ is an Abelian group. The orange arrows represent frameworks for the learning of lattice operators via the SLDA and via model selection based on data.} \label{lattice_iso}
\end{figure}

When one has strong prior information about the properties that $\psi^{\star}$ satisfies, then he can properly fix a constrained $\Omega$ and, having a lattice overparametrization of $\Omega$, he can in principle approximately compute \eqref{psiHat}. However, when strong prior information is not available, $\Omega$ may be too complex, so overfitting occurs, or the lattice $\Theta$ may be too complex, so high computational resources are needed. Either way, if one can decompose $\Theta$ into a lattice $(\mathbb{L}(\Theta),\subset)$ of subsets of $\Theta$ then he can apply an algorithm analogous to the SLDA to minimize a validation error in $(\mathbb{L}(\Theta),\subset)$ to select a subset $\hat{\Theta} \subset \Theta$, which represents a constrained class $\{\psi_{\theta}: \theta \in \hat{\Theta}\} \subset \Omega$, and then learn an operator in it. This is a specific instance of learning via model selection and is also represented in Figure \ref{lattice_iso} (see \cite{marcondes2023distribution} for a formal definition of learning via model selection).

We proposed in \cite{marcondes2023algorithm} a hierarchical SLDA in the context of the unrestricted sequential DMNN proposed in \cite{marcondes2023discrete} to represent W-operators. The class represented by these DMNN is composed of all operators that can be represented via the composition of $d$ W-operators locally defined in $W_{1},\dots,W_{d}$, which is overparametrized by the Boolean characteristic functions of the W-operators. The set of possible sequences of Boolean functions is a Boolean lattice, and hence this is a lattice overparametrization. Since this class is quite complex, it is prone to overfit the data, so we propose a SLDA to select the windows of the W-operators, what is equivalent to creating equivalence classes on the characteristic functions' domain. Each possible sequence of windows defines a subset of $\Theta$ and varying all possible windows generates a lattice $(\mathbb{L}(\Theta),\subset)$ of subsets of $\Theta$. This is an example where it is possible to learn lattice operators without strong prior information, and we refer to \cite{marcondes2023algorithm} for more details. 

We are currently working on more general methods to learn lattice operators via a hierarchical SLDA in contexts where prior information is not available.	

\section{Control, transparency and interpretability}

The lattice overparametrization paradigm for the machine learning of lattice operators has by design three important properties that modern learning methods lack in general: control, transparency and interpretability. Due to the extensive knowledge about lattice operators and the mathematical morphology toolbox, the practitioner has all the resources necessary to design $\Omega$ to fulfill its needs, so he has complete control over the class of operators. This is clear in the case of canonical DMNN in the context of set operators \cite{marcondes2023discrete}, which can represent any class of operators that can be decomposed via supremum, infimum, complement, and composition of erosions and dilations.

All the steps of the machine learning are transparent: the practitioner knows the properties of the operators in $\Omega$ since he can compute the basis of each one via $\tilde{R}_{\boldsymbol{B}}$; he knows the lattice overparametrization, which he chose; and he can trace the path of the SLDA and inspect the choices of the algorithm at each step. By monitoring the properties of $\theta$ each time it is updated, one can make sense of a possible logic that the algorithm is following. This monitoring may be that of the basis $\tilde{R}_{\boldsymbol{B}}(\theta)$ or of the values of the parameters $\theta$ in case they have semantic information.

Finally, the mathematical properties of the learned operator $\psi_{\hat{\theta}}$ are completely known, since it suffices to compute its basis $\tilde{R}_{\boldsymbol{B}}(\hat{\theta})$ from which its properties can be deduced. From these properties, it is possible to explain what the operator is doing, foresee cases in which it might not properly work and obtain insights about the relation between $X$ and $Y$.

We note that these three properties are present in a learning framework only if $\tilde{R}_{\boldsymbol{B}}$ can be computed, for otherwise, if one cannot reduce an operator to its basis, then he may not be able to deduce its properties. This is a possible bottleneck to this learning paradigm:

\vspace{0.1cm}

\textbf{(B3)} \textit{To compute $\tilde{R}_{\boldsymbol{B}}(\theta)$ for $\theta \in \Theta$}

\vspace{0.1cm}

For canonical DMNN in the context of set operators this is not a bottleneck since results in \cite{barrera1996set} allow computing the basis for each $\theta \in \theta$ (see \cite[Remark~5.2]{marcondes2023discrete} for more details). Moreover, results in \cite{jones1994basis} present general algorithms to compute the basis of many classes of set operators. Having these kinds of results for more general lattices is needed to overcome \textbf{(B3)}.
	
\section{Conclusion}

In the last decades, neural networks (NN) have been the main paradigm in image processing and its outstanding performance overshadowed mathematical morphology (MM), that has been relegated in favor of them (see the discussion in \cite{angulo2021some}). To this date, there has been no definitive method that brought MM to the deep learning era and many attempts try to insert MM into NN, as if NN were the \textit{golden standard} and MM was only a second class \textit{tool}. This last fact is completely false: in the context of lattice operators, NN do not have any advantage over MM from a theoretical perspective, and do not have the essence of MM, which control, transparency and interpretability are a part of. Indeed, needless to say, neural networks do not have, in general, any of these three properties: one does not have control over the class it represents, its learning algorithm and behavior is opaque, and its results are hardly interpretable.

Indeed, learning methods based on neural networks have been proposed in the last decades in the form of morphological neural networks \cite{davidson1990theory,dimitriadis2021advances,ritter1996introduction}. More recently, they have been studied in the context of convolutional neural networks \cite{franchi2020deep,hu2022learning}. Although these methods do not suffer from \textbf{(B2)}, since they can be efficiently trained, they are opaque and as much a black-box as usual neural networks. In special, it is not trivial to insert constraint into them to achieve \textbf{(B1)} and once they are trained it is not possible to solve \textbf{(B3)} efficiently. Therefore, MNN do not address all the bottlenecks, but could maybe be adapted to fit the paradigm proposed in this paper. This is a current line of research.

However, NN do have great advantages from a practical standpoint, since they obtain good results and can be efficiently trained, and a great part of this success appears to be due to the possibility of proper learn in an overparametrized context. The paradigm proposed in this paper asks the following question: what if we could consider overparametrization to learn, but do not lose control and semantic understanding? To this day, there is no way of doing so with neural networks, but we argued in this paper that it is possible with MM as long as bottlenecks \textbf{(B1)}, \textbf{(B2)} and \textbf{(B3)} are overcome.

Since solving \textbf{(B1)} has been the purpose of MM for decades, it is necessary to overcome only \textbf{(B2)} and \textbf{(B3)}. The latter can be done by extending results such as those in \cite{barrera1996set,jones1994basis} to general lattices, and the former can be overcome via implementations of algorithms analogous to the SLDA. We believe one should value MM and embrace aspects of NN, such as overparametrization, that can enhance MM without losing its spirit, instead of embracing NN and trying to insert MM into it.

The works in \cite{marcondes2023discrete} and \cite{marcondes2023algorithm} are proofs of concept of the paradigm we discussed in this paper that, we believe, could be the guide for research in the machine learning of mathematical morphology in the deep learning era. The potential of such a line of research is enormous, since even if the performance of these methods come only close to that of neural networks, they may be preferred since they are controllable, transparent, and interpretable by design. Nowadays, there is an increasing demand for methods with these characteristics, and we believe that MM is in a unique position to supply them. The lattice overparametrization paradigm could be a missing piece for MM to achieve its full potential within modern machine learning. 

%
%
\bibliographystyle{splncs04}
\bibliography{ref}
\end{document}